\documentclass[a4paper,10pt,twocolumn]{article}
\usepackage[top=45pt,bottom=55pt,left=57pt,right=55pt]{geometry}
\usepackage{url}
\usepackage{natbib}
\usepackage{booktabs}
\usepackage{multirow}
\usepackage{amsmath}
\usepackage{array}
\usepackage{epsfig}
\usepackage{amssymb,amsthm}
\usepackage{color}
\usepackage{xspace}

\newcommand{\cad}{{\sc Cadec}\xspace}

\begin{document}

\title{Concept Extraction to Identify Adverse Drug Reactions in Medical Forums: A Comparison of Algorithms}
\author{Alejandro Metke-Jimenez $\;\;\;\;$   Sarvnaz Karimi\\CSIRO, Australia\\Email: \texttt{\{alejandro.metke,sarvnaz.karimi\}@csiro.au}}
\date{}

\maketitle

\begin{abstract}

Social media is becoming an increasingly important source of information
to complement traditional pharmacovigilance methods. In order to
identify signals of potential adverse drug reactions, it is necessary to
first identify medical concepts in the social media text.  Most of the
existing studies use dictionary-based methods which are not evaluated
independently from the overall signal detection task. 

We compare different approaches to automatically identify and normalise
medical concepts in consumer reviews in medical forums. Specifically, we
implement several dictionary-based methods popular in the relevant
literature, as well as a method we suggest based on a state-of-the-art
machine learning method for entity recognition. MetaMap, a popular biomedical
concept extraction tool, is used as a baseline. Our evaluations were
performed in a controlled setting on a common corpus which is a
collection of medical forum posts annotated with concepts and linked to
controlled vocabularies such as MedDRA and SNOMED CT.

To our knowledge, our study is the first to systematically examine the
effect of popular concept extraction methods in the area of signal
detection for adverse reactions. We show that the choice of algorithm
or controlled vocabulary has a significant impact on concept
extraction, which will impact the overall signal detection process. We also show
that our proposed machine learning approach significantly outperforms
all the other methods in identification of both adverse reactions and
drugs, even when trained with a relatively small set of annotated text. 

\end{abstract}


\section{Introduction}

Adverse Drug Reactions (ADRs), also known as drug side effects, are a
major concern for public health, costing health care systems worldwide
millions of
dollars~\citep{hug2012costs,ehsani2006incidence,roughead2009medication}.
An ADR is an injury caused by a medication that is administered at the
recommended dosage, for recommended symptoms. The traditional
pharmacovigilance methods have shown limitations that have prompted the
search for alternative sources of information that might help identify
\textit{signals} of potential ADRs. These signals can then be used to
select which cases warrant a more thorough review. These assessments are
performed by regulatory agencies such as the Food and Drug
Administration (FDA) in the United States and the Therapeutic Goods
Administration (TGA) in Australia, and intend to establish a causal
effect between the drug and the ADR. If a causal link is found, and 
depending on the severity of the ADR, it will be added to the drug's
label or it might even trigger a removal of the drug from the market if
it is considered life-threatening.

Social media has been identified as a potential source of information
that could be used to find signals of potential
ADRs~\citep{benton2011identifying}. A public opinion survey conducted by
The Pew Research Center's Global Attitudes Project in
2009~\citep{fox2009social} showed that 61\% of American adults looked
for health information online, 41\% had read about someone else's
experience, and 30\% were actively creating new content. These numbers
give a strong indication about the growing importance of social media in
the area of health.

Several attempts at extracting ADR signals from social media have shown
promising results~\citep{benton2011identifying,
leaman2010towards,yang2012detecting,liu2013azd}. However, all of these
techniques first need to identify concepts of interest, such as mentions
of adverse effects, in the social media text which is unstructured and
noisy. Most current approaches use medical concept identification
techniques based on dictionary lookup, but do not evaluate this step
independently~\citep{metke2014evaluation}.  This step is critical
because errors can affect the subsequent stages of the signal detection
process. The problem of concept identification and normalisation ---
linking the identified concepts to their corresponding concepts in
controlled vocabularies --- has been studied extensively in the context
of Natural Language Processing (NLP) of clinical notes, but these
techniques have not been used in the context of ADR mining from social
media. The main reasons for this are the lack of publicly
available corpora with gold-standard annotations and the difficulty
in identifying specific concepts, such as adverse reactions, in lay people
language. The noisy nature of social media text makes concept identification a hard problem.
For example, in the corpus used in this work the drug Lipitor is spelled in seven 
different ways (Lipitor, Liptor, Lipitol, Lipiltor, Liptior, Lipior and Litpitor) and
it is also written using different case combinations (e.g. Lipitor, LIPITOR and lipitor).

The contributions of this paper are twofold:
\begin{enumerate}
\item Several existing concept identification and normalisation methods
    are evaluated in the domain of adverse effect discovery from social
    media, including the dictionary-based methods applied in the recent ADR mining
    literature, as well as a state-of-the-art machine learning method
    that has been used successfully in similar tasks in other domains;
    and,

\item A variety of evaluation metrics are used, and in some cases
    proposed, to compare the effectiveness of different methods,
    including the statistical analysis of performance improvement.

\end{enumerate}

\section{Background}

This section starts by clarifying the terminology that is used throughout the paper. Then, 
a brief introduction to the controlled vocabularies that we refer to in the literature and our 
experiments is given.

\subsection{Terminology Clarification}

Some of the terminology used in this paper has been used inconsistently in the 
literature. We use the terms concept and entity in free text interchangeably. 
Concept recognition and concept identification are also treated as synonyms. We refer to 
concept extraction as a process of concept identification followed by 
normalisation / mapping to controlled vocabularies.

We also note that the methods we refer to as dictionary-based are also
known as lexicon-based or lexicon lookup.

\subsection{Controlled vocabularies}

Controlled vocabularies are typically used to identify medical concepts in free
text. This section provides background on controlled vocabularies that are
commonly used in the relevant literature. Note that some of these resources are
really taxonomies or ontologies, but are used as controlled vocabularies in the
context of this paper. 

\paragraph*{MedDRA} The Medical Dictionary for Regulatory
Activities\footnote{\url{http://www.meddra.org/}} is a thesaurus of ADRs
used internationally by regulatory agencies and pharmaceutical companies
to consistently code ADR reports. 

Before MedDRA, the FDA had developed the Coding Symbols for a Thesaurus
of Adverse Reaction Terms (COSTART), which is now obsolete.

\paragraph*{CHV} The Consumer Health
Vocabulary\footnote{\url{http://www.consumerhealthvocab.org/}} provides
a list of health terms used by lay people, including frequent
misspellings. For example, it links both {\em lung tumor} and {\em lung
tumour} to {\em lung neoplasm}.

\paragraph*{SNOMED CT} The  Systematized Nomenclature of Medicine -
Clinical Terms\footnote{\url{http://www.ihtsdo.org/snomed-ct/}} is a
large ontology of medical concepts that has been recommended as the
reference terminology for clinical information systems in countries such
as Australia, the United Kingdom, Canada, and the United
States~\citep{lee2013survey}. It includes formal definitions, codes,
terms, and synonyms for more than 300,000 medical concepts. Several
versions of the ontology exist, including an international version and
several country-specific versions that extend the international version
to add local content and synonyms.

\paragraph*{UMLS} The Unified Medical Language
System\footnote{\url{http://www.nlm.nih.gov/research/umls}} is a
collection of several health and biomedical controlled vocabularies,
including MedDRA, SNOMED CT, and CHV. Terms in the controlled
vocabularies are mapped to UMLS concepts. It also provides a semantic
network that contains semantic types linked to each other through
semantic relationships. Each UMLS concept is assigned one or more
semantic types.

\paragraph*{AMT} The Australian Medicines
Terminology\footnote{\url{http://www.nehta.gov.au/our-work/clinical-terminology/australian-medicines-terminology}}
is an extension of the Australian version of SNOMED CT that provides
unique codes and accurate, standardised names that unambiguously
identify all commonly used medicines in Australia.

\section{Related Work}

Although there is a large body of literature on generic information
extraction from formal text such as news and social media, especially
Twitter, there is limited work on the specific area of ADR detection.
ADR signal detection has been studied in spontaneous reporting
systems~\citep{Bate:Evans:2009}, medical case
reports~\citep{Gurulingappa:Mateen:Toldo:2012}, and Electronic Health
Records~\citep{Friedman:2009}. A comprehensive survey of text and data
mining techniques used for ADR signal detection from several sources,
including social media, can be found
in~\citep{Karimi:Wang:Metke-Jimenez:2015}. Below, we review the most
relevant ADR extraction techniques used in social media. We also review
the state of the art in medical concept identification and normalisation
in the context of clinical notes.

\subsection{ADR Extraction from Social Media} \label{section:adrsm}

Medical forums are online sites where people discuss their health
concerns and share their experience with other patients or health
professionals. Actively mining these forums could potentially reveal
safety concerns regarding medications before regulators discover them
through more passive methods via official channels such as health
professionals.

\citet{leaman2010towards} proposed to mine patients' comments on health
related web sites, specifically
DailyStrength\footnote{\url{http://www.dailystrength.org/}}, to find
mentions of adverse drug events. They used a lexicon that combines
COSTART and a few other sources to extract ADR-related information from
text. In a preprocessing step, they break the posts into sentences,
tokenise the sentences, run a Part-of-Speech (POS) tagger, remove
stopwords, and stem the words using the Porter stemmer. Using a sliding
window approach, they match the lexicon entries with the preprocessed
text and then evaluate the matches against the manually annotated text.
Their data was annotated with ADRs, beneficial effects, indications, and
others. We evaluate a similar method without taking into account the
similarity between the tokens.

\citet{chee2011predicting} applied Na{\"i}ve Bayes and
Support Vector Machine classifiers to identify drugs that could
potentially become part of the watchlist of the US regulatory agency,
the FDA. They used patients posts on Health and Wellness Yahoo! Groups.
The text was processed to generate features for the classifiers. They
had two sets of features: all the words from the posts, and only those
words that matched their controlled vocabulary (that included MedDRA and
a list of diseases). Misspellings were not fixed. 

\citet{benton2011identifying} extracted potential
ADRs from a number of different breast cancer forums (such as
\url{breastcancer.org}) by using frequency counts of terms from a
controlled vocabulary in their corpus and then using association rule
mining to establish the relationship between the matching terms.
Association rule mining is a data mining approach popular for mining
ADRs from regulatory and administrative databases. The method by Benton
et al.  was an advancement on the approach used by Leaman et
al.~(\citeyear{leaman2010towards}), as they did not stop at just the
extraction of interesting concepts, but also proposed a method to
establish a relationship between the extracted terms. 

\citet{yang2012detecting} studied signal detection from
a medical forum called MedHelp using data mining approaches. They
extended the existing association rule mining algorithms by adding
``interestingness" and ``impressiveness" metrics. They had to find
mentions of ADRs in the text to process the forum data and calculate
confidence and leverage. To do this, they used a sliding window and the
CHV as a controlled vocabulary to match the terms.  

None of these studies
\citep{chee2011predicting,benton2011identifying,yang2012detecting}
evaluated the information extraction step on its own, which we will cover
in this study.  

\citet{liu2013azd} implemented a system called AZDrugMiner. Data was
collected using a crawler and was then post-processed by removing 
any HTML tags and extracting text for further analysis. They then used an NLP tool
called OpenNLP to break the text into sentences.  To find the relevant
parts of each sentence, for example  mentions of a drug, they used
MetaMap~\citep{Aronson:2001}, which maps text to UMLS concepts. After
this stage, they extracted relations using co-occurrence analysis. They
also used a tool called NegEx~\citep{chapman2001simple} to identify
negations in the text. This work uses MetaMap for the concept extraction
step which we use as a baseline in our work.

\citet{Sampathkumar:Chen:Luo:2014} proposed to used a machine learning
approach, Hidden Markov Model, to extract relationships between drugs
and their side effects in a medical forum called medications.com. For
their concept recognition module they relied on a dictionary-lookup
method.They created a dictionary of drug names from the drugs.com website
and a dictionary of adverse drug effects from SIDER, a resource that
lists side effect terminology. The concept recognition step was not
evaluated on its own. 

\citet{metke2014evaluation} empirically evaluated a lexicon-based
concept identification mechanism, similar to the ones reported in the
existing literature, and tested different combinations of preprocessing
techniques and controlled vocabularies using a manually annotated data
set of medical forum posts from the AskaPatient website. The results
showed that the best performing controlled vocabulary was the CHV, but
the overall performance was quite poor. Our work has a similar goal but
differs because we compare more methods, including a baseline method and
a state of the art machine learning method. Also, the data set we use is
larger and contains a wider variety of posts. The task we evaluate also
includes \textit{concept normalisation} which requires mapping the spans
that were identified to a corresponding concept in a controlled
vocabulary. Finally, we use more comprehensive metrics to
compare the relative performance of the different techniques under evaluation.

\subsection{Medical Concept Identification and Normalisation}

The problem of medical concept identification and normalisation has been
extensively studied by the clinical text mining community. Early work
 often relied on pattern matching rules ---
e.g.,~\citep{Evans:Brownlow:Hersh:1996} --- or used MetaMap as a tool to
identify concepts using the UMLS Metathesaurus ---
e.g.,~\citep{Jimeno:Jimenez-Ruiz:2008}.    

More recently, several open challenges have bolstered the research in
this area, including the i2b2 Medication Extraction
Challenge~\citep{Uzuner:Solti:Xia:2010}, ShARe/CLEF eHealth Evaluation
Lab 2013 and SemEval-2014. 

In 2010, the i2b2 medication extraction challenge was introduced as an
annotation exercise. Participating teams were given a small number of
discharge summaries (10 per person) to annotate for mentions of
medications, the way these medications were administered (dosage,
duration, frequency, and route), as well as reasons for taking the
medications. To complete this challenge some participating teams used
automated methods as well as manual reviews.
\citet{Mork:Bodenreider:2010}, for example, used a combination of
dictionary lookup (e.g., UMLS, RxTerms, DailyMed) and concept annotation
tools (e.g., MetaMap) to find the concepts. 

Task 1 of the ShARe/CLEF eHealth Evaluation Lab
2013~\citep{pradhan2013task} used the ShARe corpus, which provides a
collection of annotated, de-identified clinical reports from US
intensive care units (version 2.5 of the MIMIC II
database\footnote{\url{http://mimic.physionet.org}}).

The task was divided into two parts. The goal of part A was to identify
spans that represent {\em disorders}, defined as any text that can be
mapped to a SNOMED CT concept in the Disorder semantic group. The goal
of Part B was to map these spans to SNOMED~CT codes. Part A was
evaluated using precision, recall, and F-Score (see
Section~\ref{sec:concept_identification} for the definition of these
metrics). Evaluation for concept identification was divided into two
categories: {\em strict} and {\em relaxed}. The strict version
required that the annotations match exactly, while the relaxed version
did not. Part B was evaluated using accuracy, which was defined as the
number of pre-annotated spans with correctly generated codes divided by
the total number of  pre-annotated spans (note that in this paper this
metric is referred to as \textit{effectiveness}; see
Section~\ref{sec:concept_normalisation}).  The strict version considered
the total number of pre-annotated spans to be the total number of
entities in the gold standard. The relaxed version considered the total
to be the number of strictly correct spans generated by the system in
part A. 

All the best performing systems for part A (concept identification)
 used machine learning algorithms, including Conditional Random Fields (CRF)
and Structural Support Vector Machines (SSVM)~\citep{tang2013recognizing,
leaman2013ncbi, gung2013using}. Our work extends on the definition of this task
by increasing the number of concepts to be identified, and tailoring
it to the adverse effect signal detection area. We also target forum data
which raises its own specific challenges due to language irregularities.

    \begin{table*} [tb] \footnotesize
        \caption{\label{table:clef_semeval_a} Best reported scores in part A (concept identification) of ShARe/CLEF 2013 Task 1 and SemEval 2014 Task 7.}
        \begin{center}
	\tabcolsep 3pt
        \begin{tabular}{llccc}
        \toprule
        \textbf{Task}&\textbf{Strictness}&\textbf{Precision}&\textbf{Recall}&\textbf{F-Score}\\
	\midrule
        CLEF~\citep{tang2013recognizing}&\multirow{2}{*}{Strict}&0.800&0.706&0.750\\
        SemEval~\citep{zhang2014uth}	&&0.843&0.786& 0.813\\
	\midrule
        CLEF~\citep{tang2013recognizing}&\multirow{2}{*}{Relaxed}&0.925&0.827&0.873\\
        SemEval~\citep{zhang2014uth}&   &0.916&0.907&0.911\\
	\bottomrule
        \end{tabular}
        \end{center}
    \end{table*}

    \begin{table} [tb] \footnotesize
        \caption{\label{table:clef_semeval_b} Best reported scores in part B (SNOMED CT mapping) of ShARe/CLEF 2013 Task 1 and SemEval 2014 Task 7.}
	\tabcolsep 3pt
        \begin{center}
	\begin{tabular}{llc}
	\toprule
        \textbf{Task}&\textbf{Strictness}&\textbf{Accuracy}\\ 
	\midrule
        CLEF~\citep{leaman2013ncbi}& \multirow{2}{*}{Strict} & 0.589 \\
        SemEval~\citep{zhang2014uth}& & 0.741  \\ 
	\midrule
        CLEF~\citep{leaman2013ncbi} & \multirow{2}{*}{Relaxed} & 0.895\\
        SemEval~\citep{zhang2014uth}& & 0.873 \\
	\bottomrule
        \end{tabular}
        \end{center}
    \end{table}

Task 7 at SemEval-2014 was a
continuation of the CLEF 2013 task, but used more data for training and
introduced a new test set~\citep{pradhan2014semeval}. The best scores
obtained in these challenges are shown in
Tables~\ref{table:clef_semeval_a} and \ref{table:clef_semeval_b}.

    \begin{table} [tb] \footnotesize
        \caption{\label{table:features}Some of the common features used in machine learning approaches to disorder span identification.}
	\begin{footnotesize}
        \begin{center}
        \begin{tabular}{| l | p{0.55\linewidth} |}
            \hline
            \textbf{Feature} & \textbf{Description} \\ \hline
             Bag of words & The words that surround each token. \\ \hline
             POS tags & The part of speech tag assigned to the token.  \\ \hline
             Word shape & Indicates the shape of the token, for example, if the token is composed of only lower case letters, upper case letters, or a combination. \\ \hline
             Type of notes & The corpus includes different types of clinical notes (e.g. discharge summaries, radiology reports, etc.). This feature indicates the type of note that contains the token.\\ \hline
             Section information & Indicates the section of the note that contains the token (e.g. Past Medical History). \\ \hline
             Semantic mapping & The concept assigned to a token by an existing tool, such as MetaMap or cTAKEs.  \\ \hline
        \end{tabular}
        \end{center}
    \end{footnotesize}
    \end{table}

The main differences between the different implementations submitted to
these challenges was the selection of features used as input to the
machine learning algorithms. Table~\ref{table:features} shows some of
the common features used by different systems. Note that not all systems
reported their features.

Apart from these open challenges, a recent study
by~\citet{Ramesh:Belknap:2014} also proposed using supervised machine
learning, including Na\"ive Bayes, support vector machines, and
conditional random fields, to annotate ADR reports collected by the FDA
for drugs and adverse effects and then reviewing the annotation using
human annotators. The main goal of this study however was developing an
annotated corpus of drug reviews and therefore different to our study in
terms of the evaluations involved.  

There is another line of studies that are also referred as {\em
normalisation}, or more specifically social text normalisation, in the
natural language processing domain.  Studies such as
\citep{Hassan:Menezes:2013,Ling:Dyer:Black:2013,Chrupala:2014} propose
algorithms to restore the standard or formal form of non-standard words
that appear frequently in social media text. For example, {\em chk} and
{\em abt}, two abbreviations common in Twitter, may be normalised to
{\em check} and {\em about}. In our work, we refer to normalisation as
mapping specific medical concepts to biomedical ontologies and
controlled vocabularies, which is different to transforming a given
free text abbreviation to its formal equivalent. 


\section{Problem Formulation}

Our goal is to evaluate the concept identification and normalisation
step independently from the overall task of signal detection in free-text.
We restrict our task to focus on social media, specifically medical
forums. Apart from the challenges that this data type raises, such as
dealing with misspellings and colloquial language, we also aim to
evaluate of concept identification techniques that are widely used in
the literature to determine how well they perform in comparison to each
other.  Since in the specific application of ADR
signal detection, linking the concepts to a standard vocabulary
provides another level of knowledge that can be utilised, we also
evaluate this step which we call normalisation. In this section, we
formally describe these two parts of the task: {\em concept
identification} and {\em concept normalisation}, and their evaluation
metrics.  

\subsection{Concept Identification}
\label{sec:concept_identification}

Concept identification consists of identifying spans of text that
represent medical concepts, specifically drugs and ADRs. The latter is more 
challenging because the same medical concept can be
considered an ADR, a symptom, or a disease, depending on the context in
which it is used. We avoid dealing with this complexity and therefore
define the goal of the task to be the identification of any span of
text that could represent a drug or an ADR, disregarding the context.

\begin{figure*}[tb]
  \centering
  \includegraphics[width=0.75\textwidth]{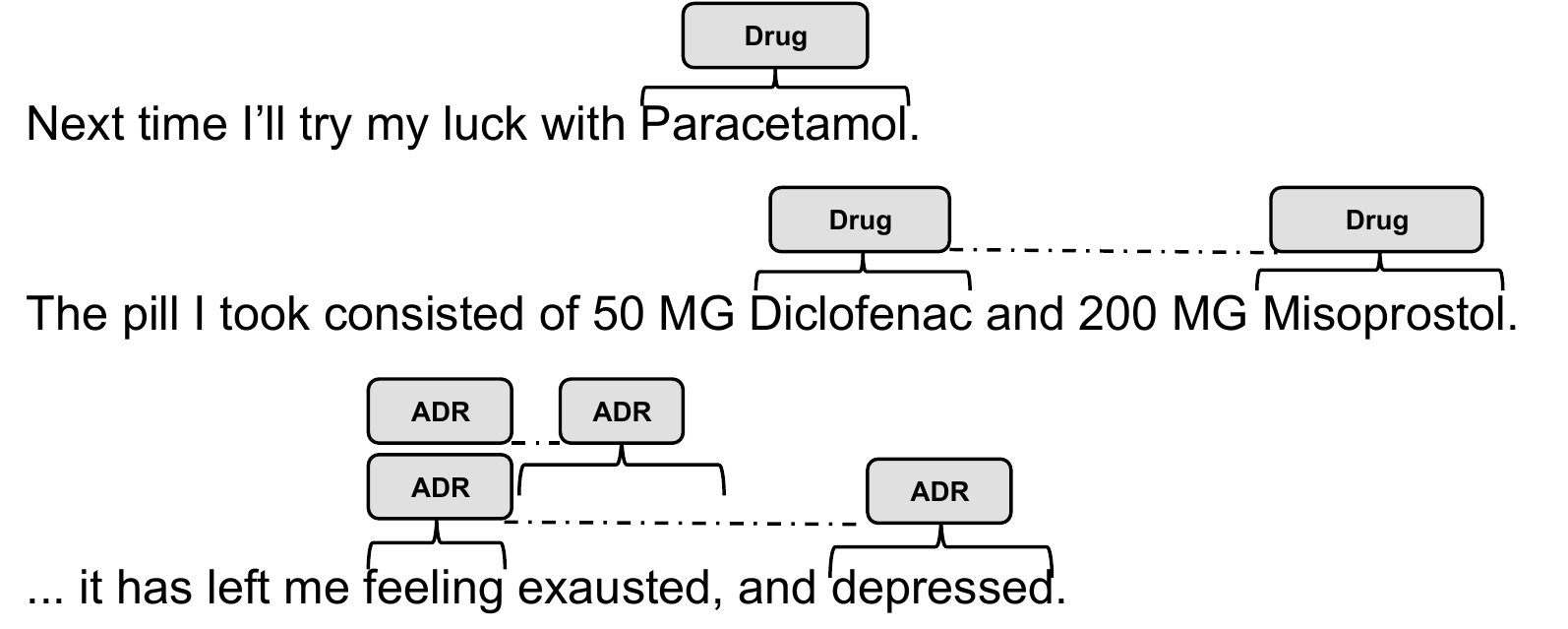}
  \caption{\label{figure:sentences}Span type examples from our dataset. From top to
  bottom: a sentence with a continuous annotation; a sentence with a
  discontinuous annotation; and a sentence with multiple discontinuous
  annotations that share a common fragment.}

\end{figure*}

Spans can be continuous or discontinuous. Spans cannot overlap each
other, except when several discontinuous spans share a common fragment.
Figure~\ref{figure:sentences} shows some examples of these different
span types. In the presence of potentially overlapping spans, the
annotators were asked to select the longest one.

Concept identification can be framed as a binary classification problem
and evaluated using precision, recall, and F-score as defined
below

\begin{center}
\begin{tabular}{>{$}l<{$}}
\text{Precision}=\frac{n_{TP}}{n_{TP}+n_{FP}},\\
\\
\text{Recall}=\frac{n_{TP}}{n_{TP}+n_{FN}},\\
\\
\text{F-Score}=2 \times \frac{precision \times recall}{precision + recall},\\
\end{tabular}
\end{center}

\noindent
where $n_{TP}$ is the number of matching spans, $n_{FP}$ is the number 
of spans reported by the system that are
not part of the gold standard, and $n_{FN}$ is the number of spans in
the gold standard that were not reported by the system. In the strict
version of the evaluation, the spans are required to match exactly. In
the relaxed version the spans only need to overlap to
be considered a positive match. In this case, however, only one to one
mappings are allowed, i.e. a span can only be mapped to one other span.

These metrics do not consider the correct classification of negative
examples~\citep{sokolova2009systematic}. In order to measure the overall
effectiveness of each system, we propose to use accuracy, which
is defined as
\begin{displaymath}
    \text{Accuracy}=\frac{n_{TP} + n_{TN}}{n_{TP}+n_{FN}+n_{FP}+n_{TN}},
\end{displaymath}
\noindent
where $n_{TN}$ is the number of spans that are not in the gold standard
that were not generated by the implementation under evaluation. Notice
that in this task, any span that is not part of the gold standard is
considered an incorrect span; negative examples are not explicitly
enumerated. Given that the total number of negative examples is
extremely large and that we are interested in comparing several
methods, the set of negative examples is defined as all the
spans that are created by all the methods under evaluation that
are not part of the gold standard.

\subsection{Concept normalisation}
\label{sec:concept_normalisation}

The normalisation step takes the spans that were identified in the
identification step and maps them to a concept in an ontology or
controlled vocabulary. For example, all three mentions of medications
{\em Pethidine}, {\em Demerol}, and {\em Meperidine} are all mapped to
one, {\em Pethidine}. This step helps to find the
links to concepts that are semantically similar or identical.  

In our setting, ADR spans are mapped to the {\em Clinical Finding} hierarchy of SNOMED
CT and in the case of drugs to a representative concept in AMT.

Concept normalisation is often evaluated using a metric
referred to as accuracy. To avoid confusion with the proposed metric for
the first part of the task, we refer to this metric as effectiveness,
which is defined as

\begin{displaymath}
    \text{Effectiveness}_{strict}=\frac{n_{TP} \cap n_{correct}}{t_g},
\end{displaymath}

\begin{displaymath}
    \text{Effectiveness}_{relaxed}=\frac{n_{TP} \cap n_{correct}}{n_{TP}},
\end{displaymath}

\noindent
where $n_{TP}$ is the number of spans that match the gold
standard exactly, $n_{correct}$ is the number of spans that were mapped to
the correct concept in the corresponding ontology, and $t_g$ is the
total number of identified concepts or spans in the gold standard.
Notice that the relaxed effectiveness metric only considers the spans
that were correctly identified in the concept identification stage,
therefore a system that performs very poorly overall can still get a
very high score on this metric.


\begin{figure*}[tb]
    \centering
    \fbox{
	\includegraphics[width=0.85\textwidth]{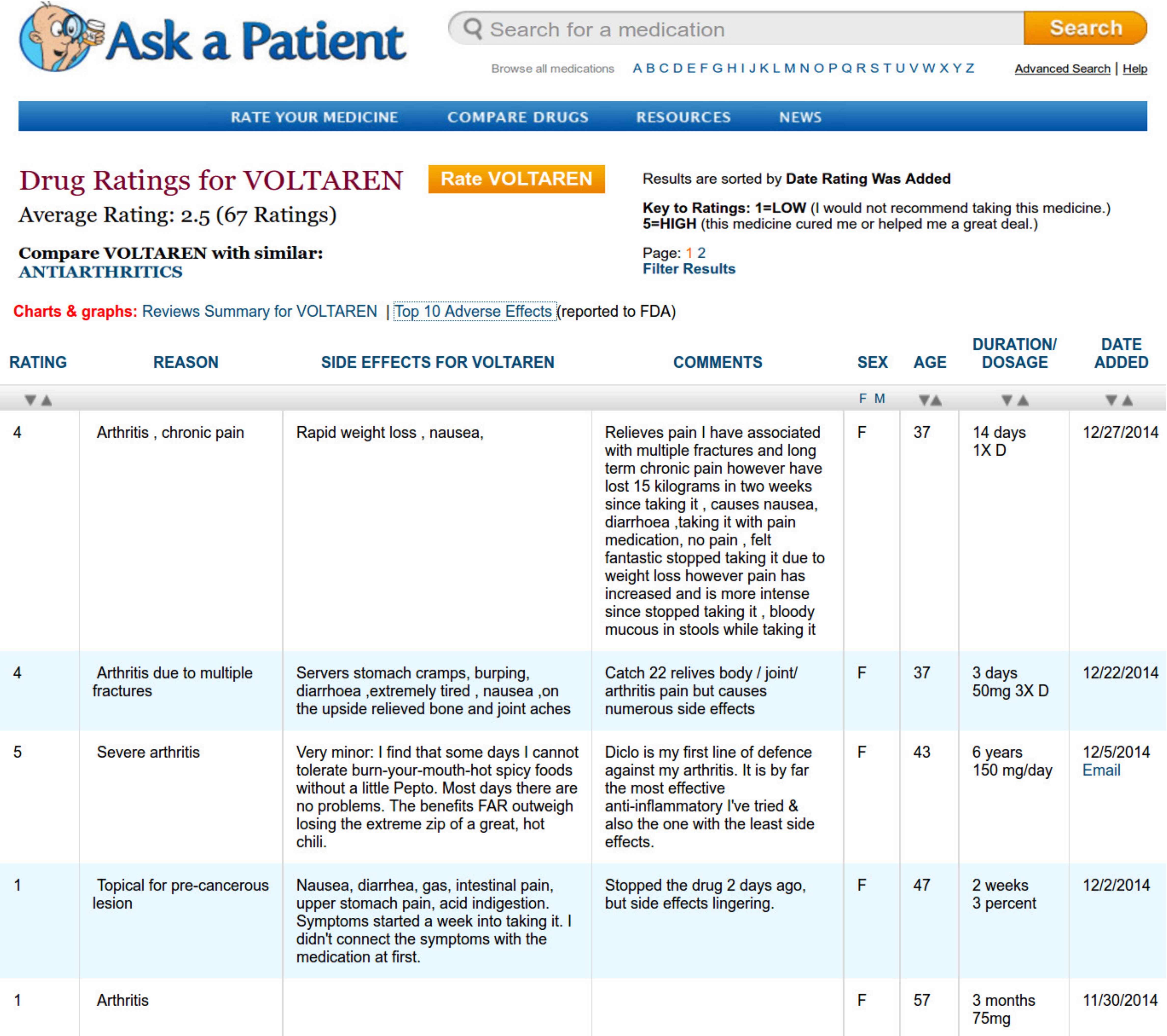}
    }
    \caption{\label{figure:snapshot}A screenshot of AskaPatient forum posts on Voltaren.}
\end{figure*}

\section{Dataset}

In our experiments, we used a publicly available annotated corpus called
CSIRO Adverse Drug Event Corpus
(\cad)\footnote{\url{http://dx.doi.org/10.4225/08/5490FA2E01A90}}. This corpus is a
collection of medical posts sourced from the medical forum
AskaPatient\footnote{\url{http://www.askapatient.com}}. The forum is
organised by drug names and allows consumers to post reviews on the
medications that they are consuming in natural language.
Figure~\ref{figure:snapshot} shows a sample from the AskaPatient website on
Voltaren. For each post shown in one row, \cad only contains two
free-text columns: side effects and comments. 

\cad includes reviews on 12 drugs, a total of 1250 forum posts. These
reviews were manually annotated with a set of tags such as drug name, and
disease name as shown in Table~\ref{table:tags}.  An expert clinical
terminologist then mapped these spans to concepts in MedDRA, SNOMED CT
and AMT. When no corresponding concept was available in the ontologies
to represent the span, the value {\em concept\_less} was assigned. A
detailed description of the corpus, including the annotation guidelines,
can be found in~\citep{Karimi:Metke-Jimenez:Kemp:2015}.

\begin{table} [tb]
    \footnotesize
    \caption{\label{table:tags}The concepts annotated in the \cad corpus.}
    \begin{center}
    \begin{tabular}{| l | p{0.67\linewidth} |}
        \hline
         \textbf{Tag} & \textbf{Description} \\ \hline
         Drug & A mention of a medicine or drug. Medicinal products and trade names are included, but not drug classes (such as NSAIDs).\\ \hline
         ADR & Mentions of adverse drug reactions clearly associated with the drug referenced by the post.\\ \hline
         Disease & A mention of a disease that is the reason for the patient taking the drug.\\ \hline 
         Symptom & A mention of a symptom that is the reason for the patient taking the drug.\\ \hline
         Finding & Any other mention of a clinical finding that does not fit into the previous categories, for example, the mention of a disease that is not the reason for the patient taking the drug.\\ \hline
    \end{tabular}
    \end{center}
\end{table}    

To develop and evaluate a machine learning approach, we divided the data
into training and testing sets, using a 70/30 split. Unlike some
previous work such as~\citep{Sampathkumar:Chen:Luo:2014}, we do not use
k-fold cross-validation to avoid potential bias that may be introduced
due to the nature of social media text~\citep{Karimi:Yin:Baum:2015}.
Table~\ref{table:dataset} shows the number of documents and span types
in each set.

\begin{table} [tb]
    \footnotesize
    \caption{\label{table:dataset}Number of documents and span types in the training and test sets.}
    \begin{center}
    \tabcolsep 3pt
    \begin{tabular}{| l | r | r | r |}
        \hline
         & \textbf{Training} & \textbf{Test} & \textbf{Total} \\ \hline
         Documents & 875 & 375 & 1250 \\ \hline
         Continuous spans & 5702 & 2350 & 8052  \\ \hline
         Discontinuous, non-overlapping spans & 57 & 37 & 94  \\ \hline
         Discontinuous, overlapping spans & 688 & 281 & 969 \\ \hline
         Total spans & 6447 & 2668 & 9115 \\ \hline
    \end{tabular}
    \end{center}
\end{table}    

\section{Methods}
\label{method}

There are existing tools that are capable of extracting medical concepts
from free text. One of these tools, MetaMap, is used as a baseline for
the evaluation. The performance is expected to be poor mainly because
MetaMap was not designed to work with social media text, which presents
several challenges such as irregularities, including misspellings,
colloquial phrases, and even novel phrases.

\subsection{Dictionary-based Approaches}

As discussed in Section~\ref{section:adrsm}, most existing approaches to
ADR mining in social media use dictionary-based techniques based on
pattern matching rules or sliding windows to identify drugs and adverse
effects in noisy text. However, these techniques have never been
evaluated independently of the overall task, nor been systematically
compared to each other under one setting. This is in part because no
standard testing set was publicly available previously.

\begin{figure*}[tb]
  \centering
  \includegraphics[width=0.75\textwidth]{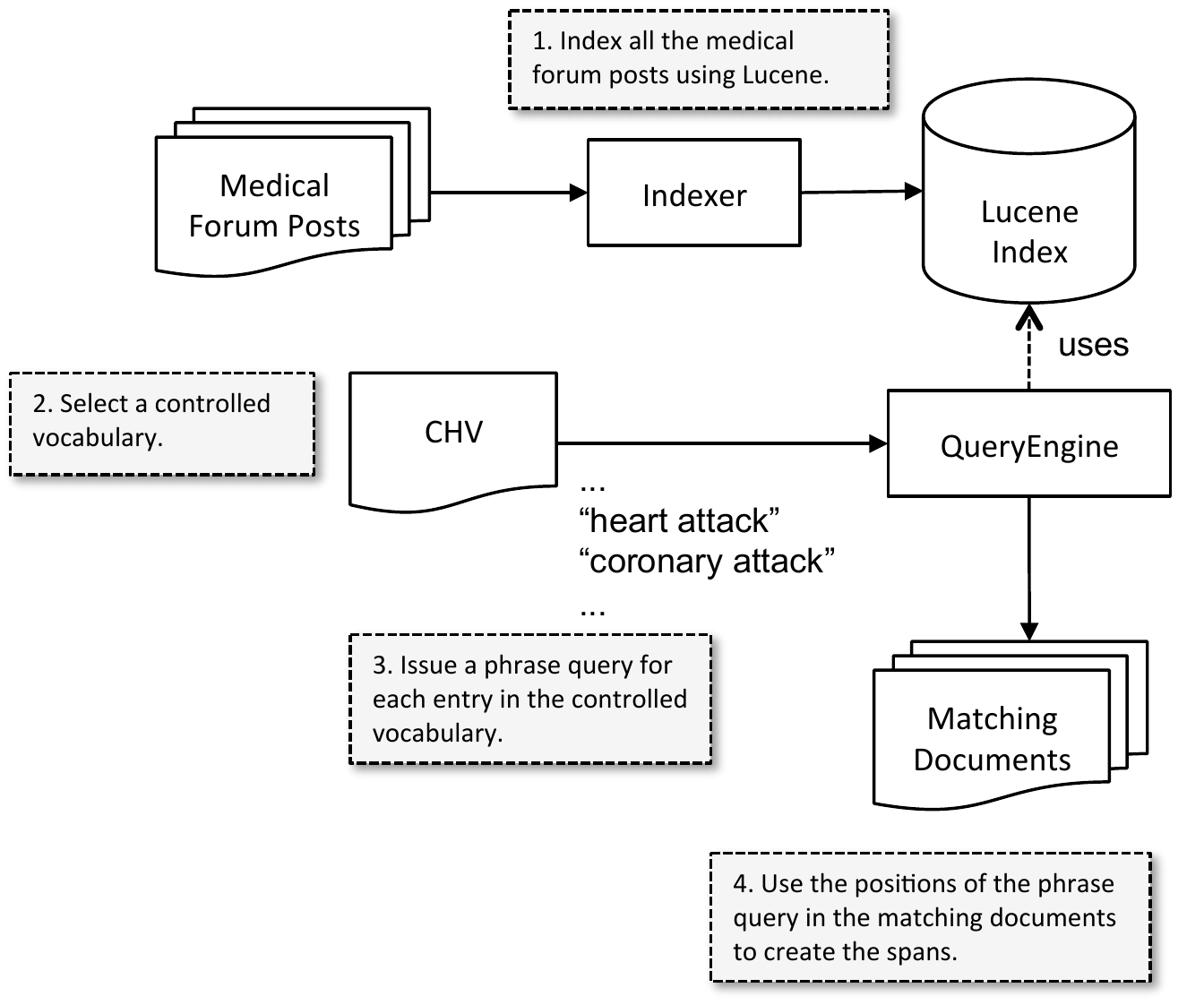}
  \caption{\label{figure:lucene}Diagram illustrating how the Lucene dictionary-based implementations work.}
\end{figure*}

We implemented a method similar to the sliding window approach
used by~\citet{yang2012detecting}, but using the Lucene search engine. The
medical forum posts were indexed and every post became a {\em document}. Then,
a controlled vocabulary was chosen and for each entry a phrase search
was executed. No stemming or stop word removal were used and tokenisation was
done using Lucene's standard tokeniser, a grammar-based tokeniser that
implements the Word Break rules from the Unicode Text Segmentation
algorithm. Any matches were transformed into spans.
Figure~\ref{figure:lucene} illustrates how this approach works when CHV
is used as the controlled vocabulary. The process is the same when replacing
other vocabularies.

Notice that the concept normalisation step is implicitly being done when
the spans are created. In the event that two different concepts match
the same identical span, the system always selects the concept with the
lexicographically greater concept id.  

\subsection{Machine Learning Approaches}

Several machine learning approaches have been used successfully to 
do entity recognition in natural language text. For example, CRF
classifiers have been used to identify medical concepts in Electronic
Health Records (EHRs); however, social media text has very different
characteristics and is typically noisier. Even though these techniques
\textit{learn} from the data, this does not necessarily mean that the
performance will be comparable.

To implement this approach, we used the CRF classifier from the Stanford
NER suite~\citep{finkel2005incorporating}\footnote{\url{http://nlp.stanford.edu/software/CRF-NER.shtml}}.
A CRF classifier takes as input different features that are derived
from the text. The features used in our implementation are listed in
Table~\ref{table:our_features}.

    \begin{table} [tb]
	\footnotesize
        \caption{\label{table:our_features}The features used in our CRF implementation.}
        \begin{center}
        \begin{tabular}{| l | p{0.67\linewidth} |}
            \hline
            \textbf{Feature} & \textbf{Description} \\ \hline
             Bag of words & The words that surround each token. \\ \hline
             N-grams & Creates features from letter n-grams (substrings of the word). In this implementation n was set to 6. \\ \hline
             Word shape & Indicates the shape of the token, for example, if the token is composed of only lower case letters, upper case letters, or a combination. \\ \hline
        \end{tabular}
        \end{center}
    \end{table}    

One of the challenges of dealing with discontinuous spans is
representing them in a format that is suitable as input to the
classifier. Continuous spans are typically represented using the
standard Begin, Inside, Outside (BIO) chunking representation common in
the most NLP applications, which assigns a B to the first token in a
span, an I to all the other contiguous tokens in the span, and an O to
all other tokens that do not belong to any span. This format does not
support the notion of discontinuous spans and several solutions have
been proposed in previous research to overcome this limitation. One of
these is to treat discontinuous spans as several continuous spans and
after classification use additional machine learning techniques to
correctly reassemble them.  Another alternative is to extend the BIO
format with additional tags to represent the discontinuous spans. The
latter approach has proved more successful in the CLEF and SemEval tasks
and therefore has been used in our implementation.

\begin{figure*}[tb]
  \centering
  \includegraphics[width=0.72\textwidth]{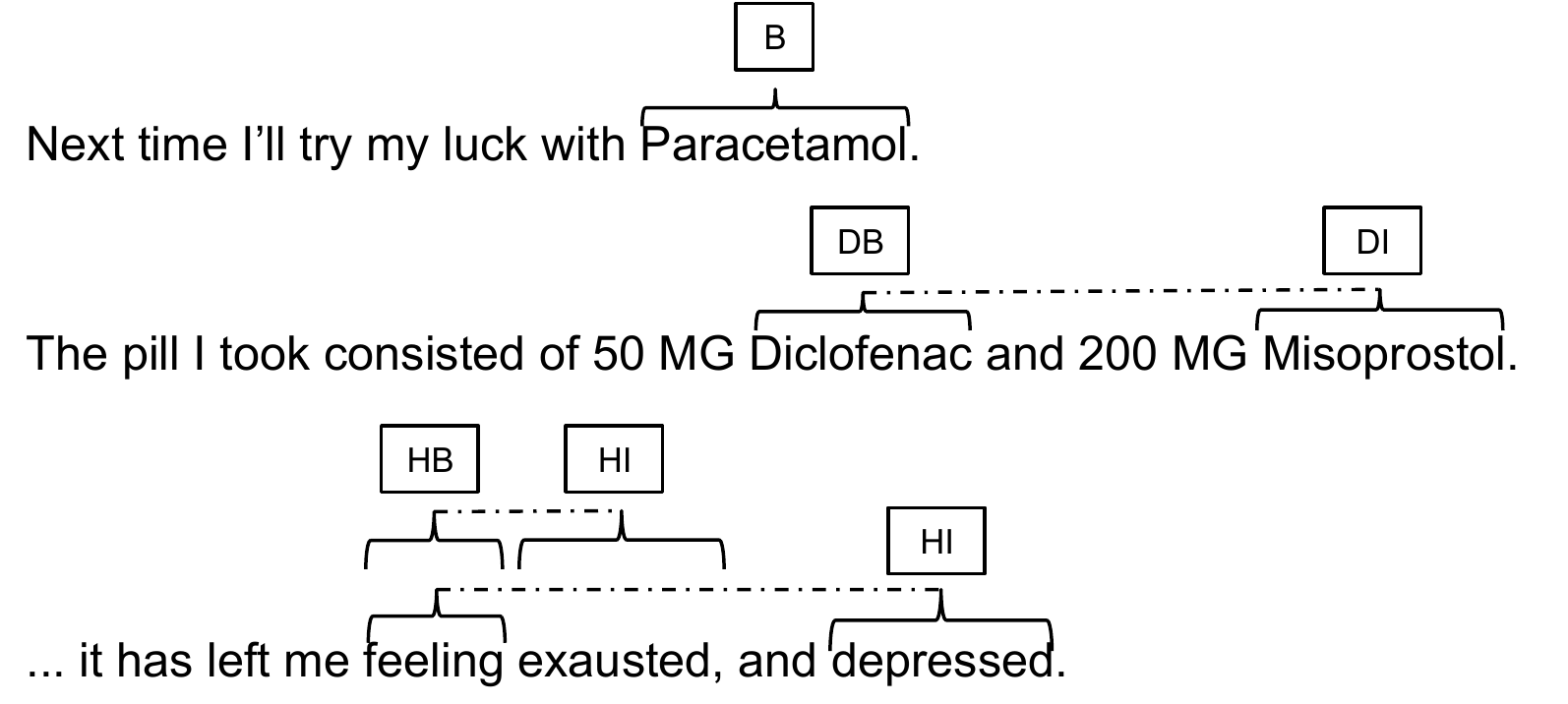}
  \caption{\label{figure:extended_bio}Examples of how annotation types
  are represented using the extended BIO format. The O annotations are
  not shown.}
\end{figure*}

With the extended BIO format, the following additional tags are
introduced: D\{B, I\} and H\{B, I\}. The first set of tags is used to
represent discontinuous, non-overlapping spans. The second set of tags
is used to represent discontinuous, overlapping spans that share one or
more tokens (the H stands for {\em Head}, as in {\em head word}).
Figure~\ref{figure:extended_bio} shows an example of these types of
annotations and how they are represented in the extended BIO format.

\begin{table} [tb]
	\footnotesize
        \caption{\label{table:roundtrip}The number of true positives (TP), false positives (FP), and false negative (FN) spans that are created from the process of transforming the ground truth spans into the extended BIO format and back. This is equivalent to having a perfect classifier.}
        \begin{center}
        \begin{tabular}{| l | r  r  r| r |}
            \hline
            \textbf{Set}& \textbf{TP} & \textbf{FP} &\textbf{FN} & \textbf{Total}\\ \hline
            \bf Training& 6325 & 122& 66 & 6513\\
            \bf Test	& 2618 & 50 & 26 & 2694\\ \hline
            \bf Total	& 8943 & 172& 92 & 9207\\ \hline
        \end{tabular}
        \end{center}
    \end{table}

Notice, however, that there is an obvious limitation with this approach:
if several discontinuous spans occur in the same sentence, then it is
impossible to represent them unambiguously. In order to determine how
this limitation might affect the performance of the CRF approach in the
social media dataset, a round trip transformation was performed, using
the gold standard annotations, that is, the gold standard was
transformed into the extended BIO format representation and then back to
the original format. This is equivalent to having a perfect CRF
classifier. Table~\ref{table:roundtrip} shows the number of correct
(TP), incorrect (FP), and spurious (FN) spans created by the round trip
process. In practice, the limitations of this format do not have a significant
impact on the overall performance. Additional techniques to deal with
ambiguous cases were not pursued and are left as future work.

One of the differences between this approach and the dictionary-based
approaches is that the CRF classifier only identifies the spans that
refer to drugs or ADRs and does not map them to the corresponding
concepts. Therefore, the second part of the task has to be implemented
independently. 

Two approaches were explored. The first one is based on a
traditional search method using the Vector Space Model (VSM). The Lucene
search engine was used for this purpose.  The target ontology was
indexed by creating a document for each term and storing the
corresponding concept id. This means that a concept with multiple
synonyms generates multiple documents in the index. In this case,
stemming and stop word removal were used. Then, the text of each span
was used to query the index. When the span included multiple tokens the
query was not required to match all of them. The top ranked concept was
assigned to the span and if the query returned no results then the span
was annotated as {\em concept\_less}.

The second approach uses Ontoserver~\citep{mcbride2012using}, a
terminology server developed at the Australian e-Health Research Centre,
that given a free-text query returns the most relevant SNOMED CT and AMT concepts.
Ontoserver uses a purpose-tuned retrieval function based on a
multi-prefix matching algorithm~\citep{sevenster2012algorithmic}. It
also supports other features such as spell checking and filtering based
on hierarchies in the ontology. We used version 2.3.0 of Ontoserver, which is publicly available at
\url{http://ontoserver.csiro.au:8080/}. The text in each span was used as a
query. The parameters were set so that all terms were not required and,
when dealing with SNOMED CT, the results were filtered so that only those concepts
that belong to the Clinical Finding hierarchy were returned. When a
query returned no results, the span was annotated as {\em concept\_less}.

A summary of all the methods that were implemented is shown in Table~\ref{table:method_summary}.

   \begin{table} [tb] 
       \footnotesize
        \caption{\label{table:method_summary}A summary of the different methods that were evaluated.}
        \begin{center}
	\tabcolsep 3pt
        \begin{tabular}{| l | p{0.67\linewidth} |}
            \hline
            \textbf{Method} & \textbf{Description} \\ \hline
             MetaMap   & The baseline method. MetaMap was used to identify and normalise the concepts in the social media text.\\ \hline
             VSM + UMLS& A dictionary-based approach based on sliding window that uses the UMLS as the underlying controlled vocabulary.\\ \hline
             VSM + CHV & A dictionary-based method based on sliding window that uses CHV, list of colloquial health terms.\\ \hline
             VSM + SCT & A dictionary-based approach based on sliding window that uses SNOMED CT as the underlying controlled vocabulary. This implementation is used to identify ADRs. \\ \hline
             VSM + AMT & A dictionary-based approach based on sliding window that uses AMT as the underlying controlled vocabulary. This implementation is used to identify drugs. \\ \hline
             CRF + VSM & A mixed approach that uses a CRF classifier to identify the concept spans and a VSM implementation to map these spans to concepts in a controlled vocabulary (SNOMED CT for ADRs and AMT for drugs). \\ \hline
             CRF + Ontoserver & A mixed approach that uses a CRF classifier to identify the concept spans and Ontoserver to map these spans to concepts in a controlled vocabulary.  \\ \hline
        \end{tabular}
        \end{center}
    \end{table}

\subsection{Statistical Significance}

To determine if the improvements obtained with any two different methods were 
statistically significant, we used  McNemar's
test~\citep{davis2012identification}. This test is applied to paired
nominal data using a $2\times2$ contingency table to determine if row
and column marginal frequencies are equal. The contingency table is
shown in Table~\ref{table:stat}, where A is the number of correct
predictions by both methods; B is the number of correct predictions by
Method 1 where Method 2 produced an incorrect prediction; C is the
number of correct predictions by method 2 where method 1 produced an
incorrect prediction; and D is the number of incorrect predictions by
both methods.

\begin{table} [tb]\footnotesize
        \caption{\label{table:stat}The contingency table used as input to McNemar's test, used to test statistical significance.}
        \begin{center}
        \begin{tabular}{ c  c  c  c }
         & & \multicolumn{2}{c}{{\em Method 2}} \\
         & & Correct & Wrong \\ \cline{3-4}
	 \multirow{2}{*}{{\em Method 1}} & \multicolumn{1}{c|}{Correct}  & \multicolumn{1}{c|}{\textbf{A}} & \multicolumn{1}{c|}{\textbf{B}} \\ \cline{3-4}
         & \multicolumn{1}{c|}{Wrong}   & \multicolumn{1}{c|}{\textbf{C}} & \multicolumn{1}{c|}{\textbf{D}} \\ \cline{3-4}
        \end{tabular}
        \end{center}
\end{table}    

\section{Results and Discussion}

The results of the concept identification task are shown in
Table~\ref{table:a_adr}. There are several noteworthy results. First,
the CRF implementation outperforms MetaMap and all the dictionary-based
implementations in all of the metrics that were considered, in both
strict and relaxed modes. Also, notice that in some cases the overall
ranking provided by the F-Score value is different from the ranking
provided by the accuracy value. In particular, when dealing with ADR
identification, the MetaMap implementation has a higher accuracy than
the VSM+UMLS implementation despite its precision, recall and F-Score
being much lower. This happens because the VSM+UMLS implementation,
despite producing more correct spans than the MetaMap implementation,
also produces many more incorrect spans.

\begin{table*} [tb]\footnotesize
\caption{\label{table:a_adr}Evaluation results of the concept
identification task, sorted by accuracy. Statistical significant
difference with the next best performing method is indicated with $\ast$ (p
\textless 0.01).}
        \begin{center}
	    \begin{tabular}{l l l cccl}
	    \toprule
            \textbf{Entities} & \textbf{Type} & \textbf{Method} & \textbf{Precision} &\textbf{Recall} & \textbf{F-Score} & \textbf{Accuracy} \\ \hline
            \multirow{10}{*}{\bf ADRs} &\multirow{5}{*}{Strict} & VSM+UMLS & 0.264 & 0.392 & 0.316 & 0.454  \\
            & & MetaMap & 0.105 & 0.080 & 0.091 & 0.485$^\ast$\\ 		
            & & VSM+CHV & 0.457 & 0.370 & 0.409 & 0.656$^\ast$\\
            & & VSM+SCT & 0.498 & 0.352 & 0.412 & 0.678$^\ast$\\
            & & CRF     & 0.644 & 0.565 & 0.602 & 0.760$^\ast$\\ \cline{2-7}	
            & \multirow{5}{*}{Relaxed}& VSM+UMLS & 0.454  & 0.674 & 0.543 & 0.635 \\ 			
            & & VSM+CHV &  0.747 & 0.605 & 0.669 & 0.807$^\ast$\\
            & & MetaMap &  0.794 & 0.605 & 0.687 & 0.822$^\ast$\\
            & & VSM+SCT &  0.818 & 0.578 & 0.677 & 0.822  \\
            & & CRF	&  0.908 & 0.797 & 0.849 & 0.909$^\ast$\\ 
	    \midrule
            \multirow{10}{*}{\bf Drugs} & \multirow{5}{*}{Strict} & VSM+UMLS  & 0.160 & 0.882 & 0.271 & 0.546 \\		
            & & VSM+AMT & 0.160 & 0.775 & 0.266 & 0.589$^\ast$\\		
            & & MetaMap & 0.022 & 0.021 & 0.021 & 0.816$^\ast$\\	
            & & VSM+CHV & 0.468 & 0.856 & 0.605 & 0.893$^\ast$\\		
            & & CRF & 0.943 & 0.840 & 0.889 & 0.980$^\ast$\\ \cline{2-7}	
           & \multirow{5}{*}{Relaxed} &  VSM+UMLS  & 0.168 & 0.923 & 0.284 & 0.554\\
            & & VSM+AMT & 0.173 & 0.837 & 0.287 & 0.601$^\ast$\\		
	    & & MetaMap & 0.145 & 0.139 & 0.142 & 0.839$^\ast$\\
	    & & VSM+CHV & 0.489 & 0.893 & 0.632 & 0.900$^\ast$\\
	    & & CRF	& 0.979 & 0.872 & 0.923 & 0.986$^\ast$\\ 
	    \bottomrule
        \end{tabular}
        \end{center}
    \end{table*}

The task of identifying drugs is considerably different from the task of
identifying ADRs because it usually involves less ambiguity. For
example, trade products usually have no synonyms and therefore limit the
number of ways a person can refer to them (this of course does not rule
out misspellings, which are common in social media). Because of this,
intuitively, this task should be easier than the task of identifying
ADRs. The results show that the CRF implementation indeed performs
better in this task that in the ADR identification task. Note also that
MetaMap obtains very low precision and recall. This is because the tool
was not designed to identify drugs. Also, most of the dictionary-based
implementations achieve good recall but low precision; this is likely
due to some of the constraints in the annotation guidelines, for
example, that indicate that drug classes should be excluded. If the drug
classes are mentioned frequently and are part of the underlying
controlled vocabularies then this will create many false positives. In
contrast, the CRF implementation is capable of identifying some of the
common drug classes that are not annotated in the training set and is
able to avoid creating false positives in most cases.

Table~\ref{table:b_adr} shows the results of the concept normalisation
task. In this case the strict metric is more relevant, because some
implementations can achieve a very high score in the relaxed version
despite having a very poor overall performance. The results show that
Ontoserver outperforms the other approaches. Overall, however, the
results are quite poor. This highlights two important aspects of the
task. First, it is inherently difficult to map colloquial language to
ontologies that contain more formal terms. Second, because in this task
the goal is to map the spans to SNOMED CT concepts, the quality of the
results when using approaches that rely on other controlled vocabularies
will depend on the quality of the mappings between those vocabularies
and SNOMED CT. For example, when using the VSM+CHV implementation, even
if the term in the text appears in CHV, if this term is mapped to an
incorrect concept in SNOMED CT, the implementation will produce an
incorrect result. Even though this issue has not been explored in depth,
some of the potential problems include mappings to concepts that are now
inactive (and therefore will never appear in the gold standard) and
mappings to concepts in other versions of SNOMED CT (for example SNOMED
US, which shares a common subset with SNOMED AU but also includes some
local concepts that will not appear in the Australian version).  

For example, in the MetaMap implementation the concept  366981002 (Pain) is
returned as the top concept for several spans and this concept has been
replaced in the current version with 22253000 (Pain). It may be possible
to automatically replace an inactive concept with the current concept
that replaced it; however, this option was not attempted and is left as
future work.

It was also expected that the different methods would perform better
when normalising drugs than when normalising ADRs. For most
implementations this turned out to be true, except for the
dictionary-based methods that are not based on AMT. These methods were
unable to normalise any concepts at all because a map between the other
controlled vocabularies and AMT does not currently exist.

    \begin{table}[tb]\footnotesize
        \caption{\label{table:b_adr}Results of the evaluation of the concept normalisation task. Baseline is MetaMap.}
        \begin{center}
	\tabcolsep 3pt
        \begin{tabular}{ lc l l c }
            \toprule
	    \textbf{Entities} && \textbf{Type} & \textbf{Method} & \textbf{Effectiveness} \\ \midrule
	    \multirow{12}{*}{\bf ADRs} && \multirow{5}{*}{Strict} 
	       &MetaMap  & 0.029\\
            && & VSM+UMLS& 0.105\\ 			
            && & VSM+CHV & 0.106\\
            && & CRF+VSM & 0.327\\
            && & VSM+SCT & 0.332\\
            && & CRF+Ontoserver & 0.376\\ \cline{3-5}	
            && \multirow{6}{*}{Relaxed}  
               & MetaMap & 0.363\\
	    && & VSM+UMLS& 0.266\\ 			
            && & VSM+CHV & 0.287\\
            && & CRF+VSM & 0.578\\
            && & VSM+SCT & 0.943\\
            && & CRF+Ontoserver & 0.666\\ 
	    \midrule
	    \multirow{12}{*}{\bf Drugs} && \multirow{5}{*}{Strict} 
               & MetaMap  & 0.000\\
            && & VSM+UMLS & 0.000\\ 			
	    && & VSM+CHV  & 0.000\\
            && & CRF+VSM  & 0.749\\
            && & CRF+Ontoserver & 0.773\\ 
	    && & VSM+AMT  & 0.758\\ \cline{3-5}
            && \multirow{6}{*}{Relaxed}  
               & MetaMap  & 0.000 \\
            && & VSM+UMLS & 0.000 \\
	    && & VSM+CHV  & 0.000 \\ 			
            && & CRF+VSM  & 0.891\\
            && & CRF+Ontoserver& 0.920\\ 
            && & VSM+AMT  & 0.978\\ 
	    \bottomrule
        \end{tabular}
        \end{center}
    \end{table}

\begin{table*} [tb]\footnotesize
\caption{\label{table:c_adr}Results of the evaluation of the full task
applied to ADRs, sorted by accuracy. Statistical significant difference
with the next best performing method is indicated with $\ast$ (p \textless
0.01).}
        \begin{center}
        \begin{tabular}{l l l l l l}
            \hline
            \textbf{Entities} & \textbf{Name} & \textbf{Precision} &\textbf{Recall} & \textbf{F-Score} & \textbf{Accuracy} \\ 
	    \toprule
            \multirow{5}{*}{\bf ADRs} & VSM+UMLS  & 0.088 & 0.104 & 0.095 & 0.363 \\
            & MetaMap   & 0.041 & 0.029 & 0.034 & 0.468$^\ast$\\
            & VSM+CHV   & 0.218 & 0.106 & 0.143 & 0.590$^\ast$\\
            & CRF+VSM   & 0.564 & 0.327 & 0.414 & 0.702$^\ast$\\
            & VSM+AMT   & 0.572 & 0.332 & 0.420 & 0.706 \\
            & CRF+Ontoserver  & 0.771 & 0.376 & 0.506 & 0.764$^\ast$ \\ 
	    \midrule
            \multirow{5}{*}{\bf Drugs} & VSM+UMLS  & 0.000  & 0.000  & 0.000  & 0.461\\
            & VSM+AMT   & 0.163 & 0.758 & 0.269 & 0.605$^\ast$  \\
            & VSM+CHV   & 0.000 & 0.000 & 0.000 & 0.814$^\ast$\\
            & MetaMap   & 0.000 & 0.000 & 0.000 & 0.814 \\
            & CRF+VSM   & 0.988 & 0.749 & 0.852 & 0.975$^\ast$ \\
            & CRF+Ontoserver & 0.988 & 0.773 & 0.867 & 0.977 \\ 
	    \bottomrule
        \end{tabular}
        \end{center}
\end{table*}

Finally, a strict evaluation of the full task was carried out, where a
span was only considered correct if it matched the gold standard span
exactly and was annotated with the same concept. This evaluation is
important because a good free text annotation system will not only need
to identify relevant spans but also annotate them correctly. The results
for the full evaluation are shown in Table~\ref{table:c_adr}. The best
performing system overall was the CRF implementation using Ontoserver
for concept normalisation.

\section{Conclusions and Future Work}

Pharmacovigilance has passed the era where it would only rely on manual
reports of potential drug adverse effects. Actively detecting signals of
adverse drug reactions through automated methods of text mining consumer
reviews is one of the emerging areas. 

We conducted an empirical evaluation of different methods to
automatically identify and normalise medical concepts in the domain of
adverse drug reaction detection in medical forums. It included several
methods commonly used in the ADR mining literature, as well as
state-of-the-art machine learning methods that have been used in other
domains. To our knowledge this is the first study to systematically
compare the most common concept identification and normalisation
approaches used in adverse effect mining from social media under a
controlled setting. This is an important step in ADR signal detection
which determines the effectiveness of automated systems in this domain.   

The experimental results showed that the CRF implementation combined
with Ontoserver outperformed all the other methods that were evaluated,
including MetaMap and the dictionary-based methods. We believe that the
availability of the new \cad corpus and the empirical results shown in
this paper will benefit other researchers working on ADR mining methods.

In the future, we plan to improve the CRF method with additional
features, specifically domain specific features that are likely to
improve recognition of ADRs in text. 

Regarding the concept normalisation task, the results showed that there
is still room for improvement. There are two avenues to explore. The
concept normalisation could also be evaluated completely independently
by using the spans in the gold standard as input. Second, to the best of
our knowledge, existing concept normalisation implementations, including
the ones implemented in this work, do not make use of the context of the
spans. We believe more advanced methods may benefit from having access
not only to the text in the span but also to the surrounding tokens and
previously identified concepts.

Regarding the evaluation, there are several ways that the current
methods could be extended. In many use cases, generating the exact
same span as in the gold standard is not relevant. However, the current
definition of a {\em relaxed} match is too loose and might not work
appropriately in certain situations. For example, when spans tend to be
long and include multiple tokens, a single token overlap constitutes a
positive relaxed match. An improvement over the relaxed matching
criteria would be to consider the extent of the match by establishing a
threshold based on either a ratio of characters or tokens that are
required for the overlap to be considered a valid match. Using a high
threshold would ensure that the systems under this relaxed evaluation
are only producing spans with minimal differences (such as including
prepositions before a noun or adjacent punctuation symbols, for
example). Several metrics that could be adapted for this scenario have
been proposed in the area of passage retrieval~\citep{wade2005passage}.

When evaluating the concept normalisation task, the current evaluation
method only considers a span to be correct if it is assigned the same
concept found in the gold standard. However, considering that the
annotations in the \cad corpus come from an ontology, if the span is 
annotated with a concept that is very close to the concept in the gold 
standard, say a parent concept, then considering the span to be 
completely wrong seems too severe.  Modifying the evaluation metric to 
consider this `semantic distance' will likely give a better sense of the 
performance of the systems under evaluation.

\section*{Acknowledgements}

AskaPatient kindly provided the data used in this study for research
purposes only.  Ethics approval for this project was obtained from the
CSIRO ethics committee, which classified the work as low risk (CSIRO
Ecosciences \#07613).

\small
\bibliographystyle{abbrvnat} 
\bibliography{ce}

\end{document}